\pdfoutput=1

\documentclass[11pt]{article}

\usepackage[dvipsnames]{xcolor}
\usepackage{ACL2023}

\usepackage{times}
\usepackage{latexsym}

\usepackage[T1]{fontenc}

\usepackage[utf8]{inputenc}

\usepackage{microtype}

\usepackage{inconsolata}

\usepackage{color}
\usepackage{graphicx}
\usepackage{paralist}
\usepackage{amsmath}
\usepackage{cleveref}
\usepackage{amssymb}
\usepackage{listings}
\usepackage{booktabs}
\usepackage{algorithm}
\usepackage{algpseudocode}
\usepackage{bbm}
\usepackage{multirow}
\usepackage{comment}
\usepackage{makecell}

\newcommand{\tuple}[7]{$($\textit{#1}, \textit{#2}, \textit{#3}, \textit{#4},
   \textit{#5}, \textit{#6}, \textit{#7}$)$}

\newcommand{\srl}{\textbf{SRLScore}}
\usepackage{enumitem}

\title{Evaluating Factual Consistency of Texts with Semantic Role Labeling}

\author{Jing Fan\thanks{~~Both authors contributed equally to this work.} \and Dennis Aumiller$^*$ \and Michael Gertz \\
  Institute of Computer Science, Heidelberg University \\
  \texttt{j.fan@stud.uni-heidelberg.de}\\
  \texttt{\{aumiller, gertz\}@informatik.uni-heidelberg.de}}

\begin{document}
\maketitle
\begin{abstract}
Automated evaluation of text generation systems has recently seen increasing attention, particularly checking whether generated text stays truthful to input sources.
Existing methods frequently rely on an evaluation using task-specific language models, which in turn allows for little interpretability of generated scores.
We introduce \srl, a reference-free evaluation metric designed with text summarization in mind. Our approach generates fact tuples constructed from Semantic Role Labels, applied to both input and summary texts.
A final factuality score is computed by an adjustable scoring mechanism, which allows for easy adaption of the method across domains.\\
Correlation with human judgments on English summarization datasets shows that \srl\ is competitive with state-of-the-art methods and exhibits stable generalization across datasets without requiring further training or hyperparameter tuning.
We experiment with an optional co-reference resolution step, but find that the performance boost is mostly outweighed by the additional compute required.
Our metric is available online at: \url{https://github.com/heyjing/SRLScore}
\end{abstract}

\section{Introduction}

One of the remaining issues that prevents productive deployments of neural text summarization systems is the low correlation of system outputs with human preferences.
Among those, \emph{factuality}, i.e., the agreement of facts in the generated summaries with those present in the input text, is not part of the general training objectives of models, which frequently leads to hallucinated facts that are detrimental to perceived system performance~\cite{ter-hoeve-etal-2020-what, fabbri-etal-2021-summeval}.
Prior work has therefore introduced metrics for automated testing of factuality in generated text~\cite{goodrich-etal-2019-assesssing,kryscinski-etal-2020-evaluating,  yuan-etal-2021-bartscore}, which allows for a more nuanced verification of model capabilities.
In particular, one of the first relevant works by \citet{goodrich-etal-2019-assesssing} introduces the idea of representing text as a series of "fact tuples", in their case as \texttt{(subject, predicate, object)} triplets.
Their method exhibits some assumptions about the underlying data, which hampers correlation with human ratings.
For example, subject or object may vary for the same sentence meaning expressed using different syntactic structures, e.g., active and passive forms.
Semantic Role Labeling (SRL), however, allows for a syntactically independent meaning representation.
Our metric, \srl, improves factuality evaluation, building on fact tuples similar to Goodrich et al. It distinguishes itself in several ways from existing approaches, though:
\begin{enumerate}[noitemsep, topsep=0.5pt]
	\item To account for a more nuanced fact representation, we employ SRL to produce abstract representations of sentences that are \emph{independent of their syntactic formulations}.
	\item Fact tuples in \srl\ are generated on the \emph{input text} instead of gold summaries; as a consequence, our method is reference-free, and may be applied for evaluation irrespective of the availability of labeled datasets.
	\item We introduce a novel weighting scheme for fact tuple comparison, 	where adjustable weights allow for user optimization.
	\item Finally, we experiment with extensions along different parts of the pipeline, including an optional co-reference resolution step and alternative similarity scoring functions.
\end{enumerate}

\begin{figure*}[t]
	\centering
	\includegraphics[width=1.0\textwidth]{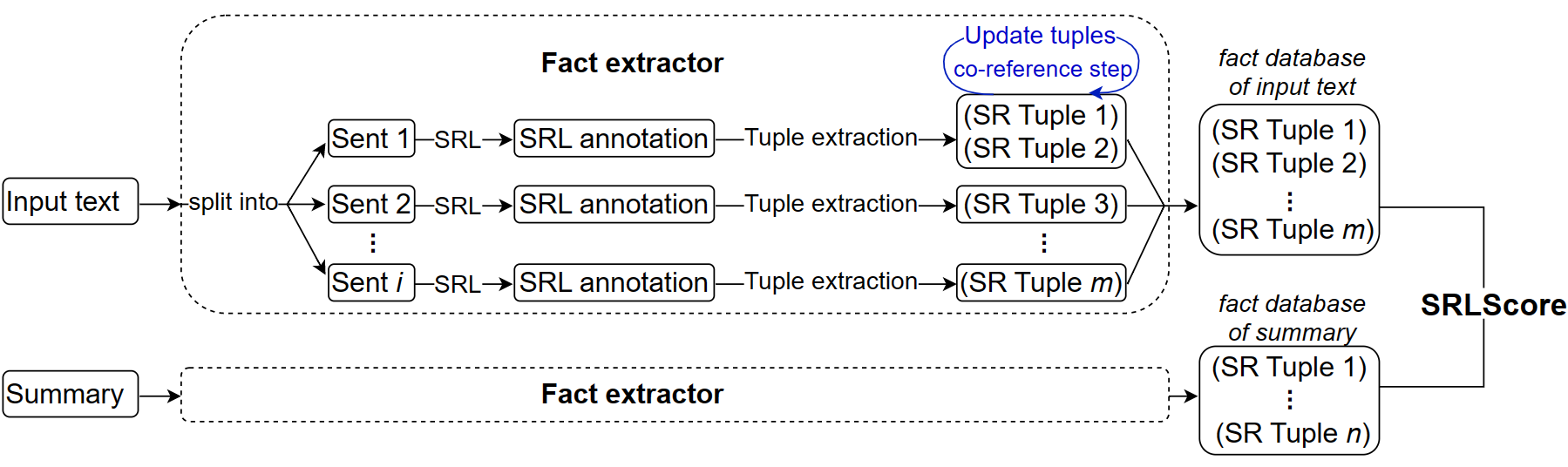}
	\caption{Visual explanation of \srl. An input text and its associated summary are transformed into a series of fact tuples (\emph{SR Tuple}) through extraction from SRL (and optional co-reference) annotations. The final factuality score is computed based on the similarity of the summary facts with fact tuples generated from the input text.}
\label{fig:overview}
\end{figure*}

Notably, \srl\ entirely relies on publicly available software components and may be used without any further domain adaption required. While our experiments are performed on English, we argue that the transfer of our approach to other languages is possible given only the existence of a language-specific tokenizer and a sufficiently good SRL tagger.
Furthermore, \srl\ offers the additional benefit of being an \emph{interpretable} metric, due to its composition on top of fact tuples. In comparison, metrics used for factuality evaluation that are based on the intermediate presentations of language models, e.g., \emph{generation perplexity}~\cite{zhang-etal-2020-bertscore,thompson-post-2020-automatic,yuan-etal-2021-bartscore}, cannot present insightful reasons \emph{why} a particular score was achieved.
Furthermore, it has been empirically demonstrated that generation-based evaluators exhibit a \emph{self-preference} of outputs generated by models similar to the factuality evaluator~\citep{fabbri-etal-2021-summeval, liu-etal-2023-geval}. This makes them a questionable choice over interpretable metrics.
We empirically show that the correlation of \srl\ with human ratings is on par with existing methods, and perform several ablations to study the impact of algorithmic choices within our pipeline.

\section{Related Work}

Automated analysis of (abstractive) summaries became more relevant in recent years, with the influx of generic summarization systems becoming available~\cite{DBLP:conf/conll/NallapatiZSGX16,see-etal-2017-get, lewis-etal-2020-bart}.
In particular, \citet{goodrich-etal-2019-assesssing} were the first to propose a reference-based estimator for factuality of generated summaries. As mentioned, their approach is based on a tuple representation of "facts" in the generated and gold summary. Fact tuples are extracted based on a weakly supervised end-to-end tagger and subsequently compared on the basis of matching arguments. Notably, no readily available implementation of their method currently exists.\\
Later work has proposed alternative metrics based on textual entailment~\cite{falke-etal-2019-ranking, mishra-etal-2021-looking} and Question Answering (QA)~\cite{wang-etal-2020-asking, durmus-etal-2020-feqa}, where agreement of answers to questions on the reference and summary are used for estimating factuality.
However, QA-based metrics require additional task-specific fine-tuning on generic datasets, which makes the adoption to new domains fairly expensive.

\noindent The only other work that to our knowledge utilizes some form of SRL-based factuality estimation is presented by \citet{fischer-etal-2022-measuring}. In comparison to \srl, their method aggregates "role buckets" at the document level, instead of creating sentence-specific fact tuples.
Empirically, their implementation has lower correlation with human ratings than compared approaches, which is contrary to our own findings.\\
\citet{li-etal-2022-just} frame factuality estimation as an in-filling task, where fact statements are withheld as masked tokens in a generated summary, and a separate model is trained to predict missing facts. Notably, this relies on the assumption that the majority of factual mistakes stems from noun phrases and entity mentions~\cite{pagnoni-etal-2021-understanding}.

\noindent An alternative body of literature has explored the possibility to exploit Language Models (LMs) directly for estimating factual consistency: Some works, such as BertScore~\cite{zhang-etal-2020-bertscore}, use LM-generated representations to generate alignments for scoring.
In comparison, PRISM~\cite{thompson-post-2020-automatic} or BARTScore~\cite{yuan-etal-2021-bartscore} directly use model perplexity as a factuality estimate. \citet{xie-etal-2021-factual-consistency} explore masking approaches, which fall somewhere between the works of \citet{li-etal-2022-just} and BARTScore; their framing of counterfactual estimation still relies on model-based likelihood scores for computation.

The majority of prior work expresses metric performance in terms of correlation with human factuality ratings. Notably, annotations exist for subsets of the popular CNN/DailyMail~\cite{hermann-etal-2015-teaching, nallapati-etal-2017-summarunner} and XSUM summarization corpora~\cite{narayan-etal-2018-dont}.
Where \citet{wang-etal-2020-asking} collect user annotations from crowd workers, \citet{fabbri-etal-2021-summeval} additionally sample expert judgments, and find that expert ratings tend to be more representative. \citet{maynez-etal-2020-faithfulness} study several aspects of summarization evaluation beyond just factuality, but do not disclose the background of annotators for evaluation.

Generally, reliably evaluating correlation of summarization metrics with human preferences is no easy task, either: \citet{deutsch-etal-2022-examining} show that system-level evaluation metrics for text summarization rarely outperform simplistic metrics, such as ROUGE~\cite{lin-2004-rouge}, to a statistically significant degree.
Partially, this can be attributed to the small number of human-annotated samples available, generally less than 1000 different instances.

\section{SRLScore}

\begin{figure}[t]
	\centerline{\includegraphics[width=0.48\textwidth]{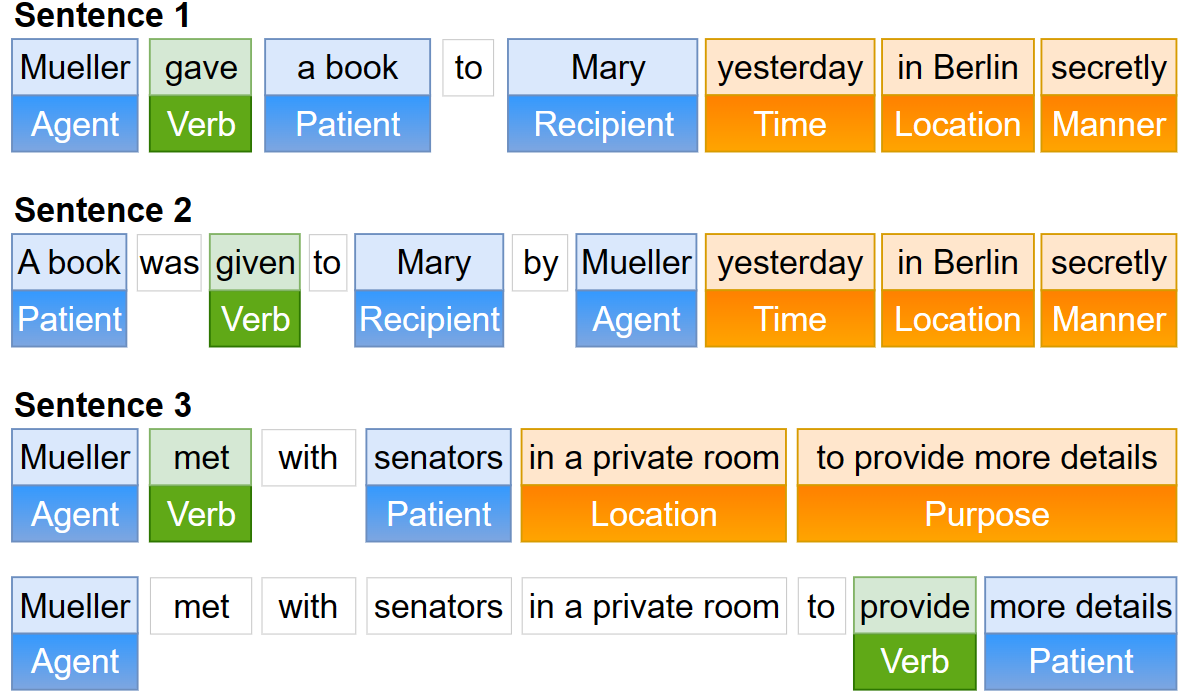}}
	\caption{Examples of semantic role label annotations. Labels may remain consistent across different syntactic forms (Sentence 1 \& 2). A single sentence can also include several relations at the same time (Sentence 3).}
	\label{srl_example}
\end{figure}

Our factual consistency metric, called \srl, is implemented as a two-stage process: first, extracting fact tuples using Semantic Role Labeling (SRL) on both the source texts and the summary texts, and then determining a factuality score based on tuple comparison. 
The measure outputs human-interpretable scores between 0 and 1, where a higher score indicates greater factual consistency of a summary text.
In this section, we detail the algorithmic choices and present an adaptive weighting scheme for computing the final factuality scores.

\subsection{Generating Fact Tuples with Semantic Role Labeling}
As \Cref{fig:overview} shows, we operate on the sentence level, primarily because existing SRL tools work well on this level of granularity~\citep{shi-lin-2019-simple,xu-etal-2021-conversational}. The goal of our fact extractor is to produce \textit{a fact database} comprised of semantic role tuples for each input text.

The primary task of SRL is to find all role-bearing constituents in a sentence and label them with their respective roles~\citep{marquez-etal-2008-semantic}. Typical semantic roles include \textit{agent}, \textit{patient/theme}, \textit{recipient}, \textit{goal}, \textit{instrument}, \textit{manner}, \textit{time}, \textit{location} and so on. From the many semantic labels available, we include seven roles based on availability in tagging schemes to construct a fact tuple: \textit{agent}, \textit{negation}, \textit{relation}, \textit{patient}, \textit{recipient}, \textit{time}, and \textit{location}.
We further note that not every sentence needs to contain \emph{all} of these roles; absent labels are represented by \textit{None} in this work. Importantly, roles reveal the semantic relations between a predicate (verb) and its arguments, which implies that one can generate several fact tuples from a single sentence, depending on the number of verbs in it. To illustrate an exemplary fact tuple, the extracted semantic tuple from sentence 1 in \Cref{srl_example} is \texttt{(Mueller, None, gave, a book, Mary, yesterday, in Berlin)}.

\subsection{Scoring Texts by Comparing Fact Tuples}\label{scoring}
Once fact tuples for both the input and summary texts are generated, the second step in our pipeline is to compute a factual accuracy score. We implement a dynamic weighting system, which crucially improves over a naive comparison, as we empirically show in \Cref{sec:dynamic}. Furthermore, we describe the drop-in replacements for exact matching during similarity computation.

\paragraph{Scoring Algorithm.} Given an input text $R$ and summary text $S$, let ${F_{R}}$ and ${F_S}$ be \emph{fact databases}, representing the semantic information contained in $R$ and $S$, respectively. Individual fact tuples are represented as an ordered list of fact arguments, e.g., $f$ =  \tuple{$agent$}{$negation$}{$relation$}{$patient$}{$recipient$}{$time$}{$location$} $ \in F$.
Particular arguments in a fact tuple are referred to by their index position, meaning \emph{agent} $ = f^0$, \emph{negation} $ = f^1$, and so on.
We further assume that there exists a scoring function that expresses the \emph{factual support of summary tuple $f_s$, given an input tuple $f_r$}, denoted as $S(f_s|f_r)$.
To obtain a factuality score, we attempt to extract the best match $\hat{f_r} \in F_R$ for each summary fact $f_s \in F_s$ where $\hat{f_r}$ maximizes the support score $S(f_s|\hat{f_r}$). 
Importantly, we differ from, e.g., \citet{goodrich-etal-2019-assesssing}, by considering the entirety of $F_R$, instead of subsets that match both the agent and relation of the fact tuple.
The factual accuracy is then the average across all maximized tuple scores in $F_S$.
With that, \srl\ is defined as:
\begin{equation}
\srl(R, S) := \frac{1}{|F_S|} \sum_{f_s \in F_s} \max_{f_r \in F_R} S(f_s|f_r)
\end{equation}
The final part of this scoring system is the computation of factual support $S(f_s|f_r)$. Tuples are scored by comparing the corresponding attributes of each tuple, formally:
\begin{equation} \label{consistency_score}
S(f_{s}|f_{r}) := \sum_i\mathbbm{1}_{f_s^i \neq  None}\cdot sim({f_s^i},{f_r^i})\cdot{w_i},
\end{equation}
where the summation over $i$ addresses all attributes of the fact tuples, $\mathbbm{1}_{f_s^i \neq  None}$ represents an indicator function considering only non-empty arguments $f^i_s$ (zero otherwise), and 
$w_i$ assigns static weights to arguments in position $i$. Generally, it should be assumed that the weights allow for a maximum factuality score of 1, i.e., $\sum_i w_i = 1$. Finally, $sim(f^i_s, f^i_r)$ is the pairwise argument similarity of $f^i_s$ and $f^i_r$. We consider different similarity metrics, as described in the following paragraphs.

\paragraph{Dynamic Weighting System.} 
The generic weighting in \Cref{consistency_score} does not necessarily apply to the particular case of evaluating factual consistency in summarization, since a summary is still factually correct even if it leaves out particular aspects (e.g., dropping the date of an event), which were present in the input text.
With static weights, however, absent arguments are still contributing to the scoring of the tuple $f_s$, which means that leaving arguments out might potentially be considered as a penalization of factuality.
To address this issue, we introduce a weight re-normalization factor, ${W_{norm}}$,
that distributes the static weights $w_i$ across only those attributes that are present in the current summary fact.
In particular, this also increases penalties for actual mistakes over simple fact omission.
The weight normalization is defined as follows:
\begin{equation} \label{weight_normalizer}
{W_{norm}} := \frac{1}{\sum\limits_i\mathbbm{1}_{f_s^i \neq  None}\cdot{w_i}} 
\end{equation}

With re-normalization enabled, we replace the existing computation of $S(f_s|f_r)$ by the product $W_{norm} \cdot S(f_s|f_r)$.

\paragraph{String Similarity Methods.}\label{sec:simi} We experiment with different methods to calculate the pairwise similarity $sim(f^i_s, f^i_r)$: exact matching (in line with prior work), but also approximate matching functions, such as word vector similarity\footnote{We use spaCy's vector similarity, see \url{https://SpaCy.io/usage/linguistic-features\#vectors-similarity}, last accessed: 2023-03-06.} and ROUGE-1 precision \citep{lin-2004-rouge}. Computation of similarity with vectors and ROUGE each have their own respective strengths. Word vectors offer the highest flexibility in terms of recognizing argument similarity, enabling semantic comparison instead of purely syntactic equivalence. ROUGE-1 similarity does not offer the same level of flexibility in terms of matching, but shines with its comparatively faster computation, while still recognizing partial matches.

\subsection{Improved Surface Form Invariance with Co-reference Resolution}

\begin{figure*}[ht]
	\centerline{\includegraphics[width=0.95\textwidth]{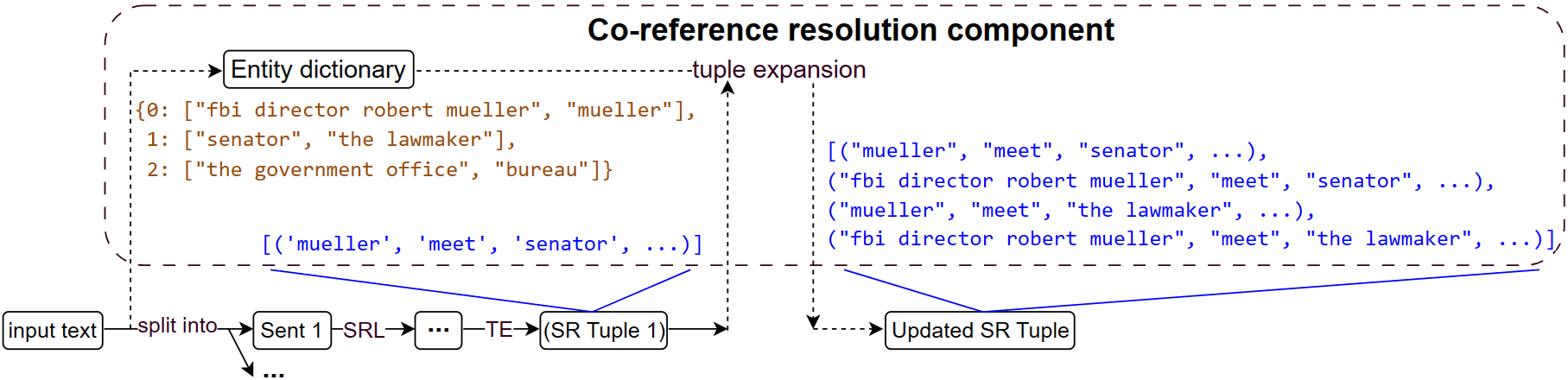}}
	\caption{Example of the tuple expansion step through co-reference resolution. In addition to the original SR tuple, we add tuples with all possible permutations of the surface forms of mentioned entities.}
	\label{coref}
\end{figure*}

In light of the fact that sentence-level SRL extraction misses co-references of the same entity across the texts, we integrate an optional component that takes co-reference resolution into account during the tuple generation.
Concretely, we employ an off-the-shelf co-reference resolution tool~\cite{lee-etal-2017-end} to identify and store all reference clusters in an external \emph{entity dictionary}. There, all linguistic expressions that refer to the same entity will be grouped together, which allows for later disambiguation.
As shown in Figure \ref{coref}, if an extracted semantic role tuple contains co-references, a single fact tuple will be \textit{expanded} into multiple tuples, representing the Cartesian product over all synonymous entity surface forms.\\
The key idea here is to enable a better matching of potential facts across input texts and summaries, effectively increasing the recall of matches. The disadvantage is that this directly affects the runtime of our method by a strong factor, since the additional tuples in $F_S$ and $F_R$ will undoubtedly increase the number of comparisons.

\section{Experiments}

We empirically demonstrate the performance of our method through a number of experiments on two popular datasets for factual consistency evaluation, which are covered in this section. We further share implementation details and the choices for extracting SRL tuples and extracting co-reference clusters. In addition to the experimental analysis, we also study the behavior of \srl\ through a number of ablation experiments and a brief error analysis.

\subsection{Evaluation Datasets}
\paragraph{QAGS \citep{wang-etal-2020-asking}.} The dataset comprises of two separate splits: the first contains 235 instances collected from the test split of CNN/DailyMail \citep{DBLP:conf/conll/NallapatiZSGX16}, where each instance contains a source article and a model-generated summary using the bottom-up approach by \citet{gehrmann-etal-2018-bottom}. A secondary set contains 239 further instances from the test split of XSUM~\citep{narayan-etal-2018-dont}, with generated summaries sampled from BART~\citep{lewis-etal-2020-bart}. 
\paragraph{SummEval \citep{fabbri-etal-2021-summeval}.} It includes synthetic summaries from 16 different abstractive and extractive models of 100 randomly selected articles from the test split of CNN/DailyMail. Unlike QAGS, which collected annotations from MTurk\footnote{\url{https://www.mturk.com/}, last accessed: 2023-03-06.}, each SummEval sample was evaluated by five crowd-sourced annotators and three experts. For each summary, judges were asked to evaluate the coherence, consistency, fluency and relevance. For our evaluation, we use the expert ratings with regard to factual consistency as the gold score, based on the recommendation by \citet{fabbri-etal-2021-summeval}.

\subsection{Evaluation Metrics and Significance} 
In line with prior work, we evaluate metrics by computing Pearson correlation (denoted as $\rho$) and Spearman correlation (denoted as $s$) between model predictions and human reference ratings.
Given the limited size of all considered evaluation datasets, we further test results for significance using permutation tests~\cite{riezler-maxwell-2005-pitfalls, deutsch-etal-2021-statistical}, following the recommendation of \citet{dror-etal-2018-hitchhikers}.
In all tables, \textsuperscript{\dag} denotes a significance level of 0.05 ($p < 0.05$) and \textsuperscript{\ddag} a level of 0.01 ($p < 0.01$). When testing significance against several systems, we further apply Bonferroni correction of significance levels~\cite{dunn-1961-multiple}.

\begin{table*}[ht]
	\centering
	
	\setlength{\tabcolsep}{10pt} 
	\renewcommand{\arraystretch}{0.93} 
	
	\resizebox{1.0\textwidth}{!}{%
	\begin{tabular}{p{0.22\linewidth}  p{0.08\linewidth}  p{0.08\linewidth}  p{0.08\linewidth} p{0.08\linewidth}p{0.08\linewidth} p{0.08\linewidth}p{0.05\linewidth}}
		\toprule
		{\multirow{2}{*}{\centering \textbf{Metrics}}} & \multicolumn{2}{l}{\centering \textbf{QAGS-CNN/DM}} & \multicolumn{2}{l}{\centering \textbf{QAGS-XSUM}} & \multicolumn{2}{l}{\centering \textbf{SummEval}} & \textbf{Avg.}  \\  		 \addlinespace[-2pt]
		\cmidrule(lr){2-3} \cmidrule(lr){4-5} \cmidrule(lr){6-7}
		{} & \hspace*{0.65em}$\rho$ & \hspace*{0.75em}\textit{s}& \hspace*{0.65em}$\rho$ & \hspace*{0.75em}\textit{s} & \hspace*{0.65em}$\rho$ & \hspace*{0.75em}\textit{s} & \hspace*{0.65em}$\rho$ \\  \addlinespace[-2pt]
		\midrule
		{ROUGE-1 (F1)} & 0.34 & 0.32 & \hspace*{-0.75em}$-$0.01 & \hspace*{-0.75em}$-$0.05 & 0.13 & 0.14 & 0.15 \\
		{BLEU} &0.13&0.33&0.08&0.03&0.09&0.14&0.10\\
		{METEOR}&0.33&0.36&0.06&0.01&0.12&0.14 &0.17\\ \addlinespace[-2pt]
		
		\midrule
		{BARTScore} &0.65&0.57&0.00&0.02&0.27&0.26&0.31\\
		{BARTScore\textsubscript{cnn}}&\textbf{0.73}&\textbf{0.68}&0.19&0.18&0.35&0.32&0.42 \\
		{BARTScore\textsubscript{cnn+para}}&0.69&0.62&0.07&0.07&0.42&\textbf{0.37}&0.39 \\
		CoCo\textsubscript{span} & 0.64 & 0.55 & 0.22 & 0.20 & 0.40 & 0.35 & 0.42 \\
		CoCo\textsubscript{sent} & 0.68 & 0.59 & 0.16 & 0.14 & 0.39 & 0.35 & 0.41\\
		ClozE-R\textsubscript{en\_core\_web\_trf}$^*$ & 0.66 & \textcolor{white}{-} - & 0.32 & \textcolor{white}{-} - & 0.47 & \textcolor{white}{-} - & \textbf{0.48}  \\
		ClozE-R\textsubscript{confidence}$^*$ & 0.65 & \textcolor{white}{-} - & 0.29 & \textcolor{white}{-} - & \textbf{0.48} & \textcolor{white}{-} - & 0.47 \\\addlinespace[-2pt]
		
		\midrule
		{\centering SRLScore\textsubscript{base}}   &0.67 & 0.59 & 0.20 & 0.18 & 0.43 & 0.33 & 0.43\\
		{\centering SRLScore\textsubscript{coref}} &0.65&0.58&{{0.27}} &{{0.26}}&0.43 &0.32 & 0.45\\
		{\centering SRLScore\textsubscript{coref-optimized}} & \textcolor{white}{-} - & \textcolor{white}{-} - &\textbf{0.33}&\textbf{0.33} & \textcolor{white}{-} - & \textcolor{white}{-} - & \textcolor{white}{-} - \\                                                                                                                
		\bottomrule
	\end{tabular}
}
	\caption{Pearson ($\rho$) and Spearman ($s$) correlation of metrics with human ratings on the evaluated datasets. Bold scores indicate highest absolute values. For \srl\ variants, we report highest scores across all similarity functions. No significant differences were found between the correlation scores of factuality-specific metrics. \\$^*$: results were taken from the respective paper, as there is no existing code to reproduce their results as of now.}
	\label{compare_SOTA}
\end{table*}

\subsection{Implementation} 
We use AllenNLP \cite{gardner-etal-2018-allennlp}, specifically version 2.1.0, to extract semantic role labels. AllenNLP implements a BERT-based SRL tagger \citep{shi-lin-2019-simple}, with some modifications. The output of AllenNLP uses PropBank convention \citep{palmer-etal-2005-proposition, propbank, pradhan-etal-2022-propbank}, which lists for each verb its permitted role labels using numbered arguments (\emph{ARG0, ARG1, ...}) instead of names, due to the difficulty of providing a small, predefined list of semantic roles that is sufficient for all verbs.
Since numbered arguments are meant to have a verb-specific meaning \citep{yi-etal-2007-can}, this implies that our mapping between numbered arguments and semantic roles may not always be consistent.
The exact mapping used in our experiments is detailed in \Cref{app:mapping}.
For co-reference, we similarly use the model provided by AllenNLP~\cite{lee-etal-2017-end}, which matches the output format of the SRL tagger.\\
All experiments were carried out on a system with an Intel Xeon Silver 4210 CPU, two TITAN RTX GPUs (24 GB GPU VRAM each) and 64 GB of main memory. We run inference for the SRL model and co-reference component on separate GPUs.\\
We report scores of all system and baseline variants across a single random seed only. Since we are comparing provided "plug-and-play" metrics, it is reasonable to assume that these are the primary choice for others evaluating their own datasets.
Particularly for \srl, we further note that due to the system design, no fine-tuning or training is necessary.
The only parameters varied during the experiments are thus the argument weights, which we describe in the following section.

\subsection{System Variants}
We compare with a number of generic automatic evaluation metrics, including BLEU~\citep{papineni-etal-2002-bleu}, ROUGE~\citep{lin-2004-rouge},  and METEOR~\citep{banerjee-lavie-2005-meteor}. Besides, we also consider several metrics specifically developed for factuality estimation, which have reported prior state-of-the-art correlation. Wherever possible, we reproduce scores with the official scripts provided by authors. Comparison is done with three variants of BARTScore~\citep{yuan-etal-2021-bartscore}, two variants of CoCo~\citep{xie-etal-2021-factual-consistency}, and two variants of ClozE~\citep{li-etal-2022-just}. For more details on reproducibility, see \Cref{app:coco}.
We chose each variant such that the highest self-reported scores of each paper on all evaluated datasets are considered.

\noindent For our own method, SRLScore\textsubscript{base} represents a default setting, assigning equal weights $w_i = \frac{1}{7}$ to all attributes \sloppy(\emph{agent, negation, relation, patient, recipient, time, location}); the respective similarity function (exact match, spaCy vector, or ROUGE similarity) is chosen to maximize dataset-specific performance (see results of~\Cref{extend_tuples}).
SRLScore\textsubscript{coref} uses the same weights, with co-reference enabled.
We further provide model ablations to test various specifications of our models.
As we could not find a implementation based on the original tuple extraction approach by \citet{goodrich-etal-2019-assesssing}, we introduce SRLScore\textsubscript{openie} and SRLScore\textsubscript{goodrich} as approximations of their method. Here, fact tuples are reduced to \texttt{(agent, relation, patient)} triplets (with equal weights $w_i = \frac{1}{3}$).
We note that this is not a true equivalence to the original method, although "[i]n most English sentences the subject is the agent" \citep{article}; in reality, a broader variety of roles in the subject position may be encountered. The same applies for our mapping between  \emph{object} and the \emph{patient} role.
However, by using the same upstream labeling tool (i.e., the SRL model provided by AllenAI), we may more accurately compare the algorithmic scoring methods, independent of the annotation accuracy.
We argue that our SRL-based modeling of relationship triplets allows for a better generalization beyond Wikipedia, which Goodrich et al.~were using in their own experiments.

The difference of SRLScore\textsubscript{openie} and SRLScore\textsubscript{goodrich} lies in the implemented scoring function, where the OpenIE variant employs our own scoring algorithm, SRLScore\textsubscript{goodrich} uses the preliminary filtering step defined in \citet{goodrich-etal-2019-assesssing}. We do not apply a co-reference system in either one of the two ablation settings.
Finally, SRLScore\textsubscript{coref-optimized} illustrates the possibility of adapting our method to a particular dataset. For this variant, we optimize available hyperparameters (weights, scoring function, co-reference) in order to obtain the highest possible scores.

\subsection{Main Results}

The central evaluation results with recommended default settings are shown in \Cref{compare_SOTA}.
In almost all cases, specialized factuality metrics show higher correlation than generic summarization evaluation metrics (ROUGE-1, BLEU and METEOR).
Notably, despite the high increase in absolute scores, we do not always detect a significant level of improvement between factuality-specific metrics and generic metrics, particularly on QAGS-XSUM; 
we will discuss further implications of this in more detail later.
When testing our own method, SRLScore\textsubscript{base}, against generic metrics, we find strongly significant improvements only for Pearson correlation of QAGS-CNN/DM and SummEval, as well as Spearman correlation on SummEval ($p < 0.01$, with Bonferroni correction).\\
It should be further noted that BARTScore\textsubscript{cnn} and CoCo results use BART models~\citep{lewis-etal-2020-bart} that were fine-tuned on the CNN/DailyMail corpus (respectively a variant fine-tuned on XSUM for CoCo on QAGS-XSUM); this may shift the results in favor of these methods for the particular dataset. In comparison, \srl\ does not make such assumptions, which may indicate a potentially stronger generalization to unseen datasets.

\noindent The results in \Cref{compare_SOTA} also show that there are no significant differences between any of the factuality-specific metrics (\srl, BARTScore, and CoCo), particularly after applying Bonferroni correction for the comparison against several methods.
These insights open up discussions about the current claims of "state-of-the-art" performance, which may not be easily distinguished on the current evaluation datasets.
We admit that there is likely no trivial solution to this (besides further annotations), as the main problem seems to stem from the high variance on small sample sizes.

\subsection{Ablation Study}

\begin{table}[t]
	\centering
	\begin{tabular}{p{0.14\linewidth} p{0.1\linewidth}  p{0.05\linewidth}  p{0.05\linewidth}  p{0.05\linewidth} p{0.05\linewidth}p{0.05\linewidth} p{0.05\linewidth}} \toprule
		\multicolumn{2}{c}{\multirow{2}{*}{\centering \textbf{\footnotesize{Metrics}}}} & \multicolumn{2}{l}{\centering \textbf{\footnotesize{QCNNDM}}} & \multicolumn{2}{l}{\centering \textbf{\footnotesize{QXSUM}}} & \multicolumn{2}{l}{\centering \textbf{\footnotesize{SummE}}}  \\  \addlinespace[-3pt]	
		\cmidrule(lr){3-4} \cmidrule(lr){5-6} \cmidrule(lr){7-8}
		\multicolumn{2}{c}{} & \hspace*{0.5em}\textit{\footnotesize{$\rho$}} & \hspace*{0.5em}\textit{\footnotesize{s}} & \hspace*{0.5em}\textit{\footnotesize{$\rho$}}& \hspace*{0.5em}\textit{\footnotesize{s}} & \hspace*{0.5em}\textit{\footnotesize{$\rho$}}& \hspace*{0.5em}\textit{\footnotesize{s}}\\ \addlinespace[-3pt]
		
		\midrule
		\multirow{3}{*}{\centering \footnotesize{\makecell{SRLScore \\ \textsubscript{openie}}}}     & \footnotesize{Exact}  &\footnotesize{0.59}             &\footnotesize{0.51}                  &\footnotesize{0.09}              &\footnotesize{0.09}           &\footnotesize{0.34}                                &\footnotesize{0.28}\\ \addlinespace[-3pt]
		& \footnotesize{ROUGE} &\footnotesize{0.62}             &\footnotesize{0.56}                  &\footnotesize{0.07}              &\footnotesize{0.07}           &\footnotesize{{0.41}}             &\footnotesize{0.32}\\ \addlinespace[-3pt]
		& \footnotesize{SpaCy} &\footnotesize{0.59}             &\footnotesize{0.53}                  &\footnotesize{0.13}              &\footnotesize{0.10}             &\footnotesize{0.37}                                &\footnotesize{0.32}\\ \addlinespace[-3pt]
		
		\midrule
		\multirow{3}{*}{\centering \footnotesize{\makecell{SRLScore\\\textsubscript{base}}}}  & \footnotesize{Exact} &\footnotesize{0.61} & \footnotesize{0.54} & \footnotesize{0.14} & \footnotesize{0.15}  &\footnotesize{0.37\textsuperscript{\dag}}     &\footnotesize{0.31\textsuperscript{\ddag}} \\   \addlinespace[-3pt]
		& \footnotesize{ROUGE}& \footnotesize{\textbf{{0.67}}} & \footnotesize{\textbf{{0.59}}} & \footnotesize{0.15\textsuperscript{\dag}}  & \footnotesize{0.13} & \footnotesize{\textbf{{0.43}}\textsuperscript{\dag}} & \footnotesize{0.33}\\    \addlinespace[-3pt]
		& \footnotesize{SpaCy} & \footnotesize{0.63} & \footnotesize{0.55}                             & \footnotesize{\textbf{0.20}}  & \footnotesize{\textbf{0.18}}& \footnotesize{0.40\textsuperscript{\dag}} & \footnotesize{\textbf{0.34}\textsuperscript{\dag}}  \\    \addlinespace[-3pt]
		
		\bottomrule
	\end{tabular}
	\caption{Comparison of \srl\ with a simplified triplet representation (SRLScore\textsubscript{openie}). Extending the fact tuples strictly improves correlation with human ratings across all similarity functions. Significance markers indicate improvements over the same similarity function of the \textsubscript{openie} variant.}
	\label{extend_tuples}
\end{table}

Given the limited expressiveness of the generic result evaluation, we perform a series of ablation studies on \srl, to support the individual algorithmic choices made in our method.

\paragraph{Extending Tuple Attributes.}
We investigate the assumption that semantic representations of sentences are usually far more complicated than the simplistic view of (\textit{agent}, \textit{relation}, \textit{patient}) triplets, and the fact that errors may involve further roles. To this end, we compared SRLScore\textsubscript{openie}, using a triplet representation, against SRLScore\textsubscript{base} with seven roles. The results in \Cref{extend_tuples} confirm that extending tuples to cover more semantic roles is effective across datasets and metrics; SRLScore\textsubscript{base} scores consistently better than SRLScore\textsubscript{openie}, with significant improvements primarily on SummEval (the largest considered dataset).

\paragraph{Performance of Similarity Functions.}
Also seen in \Cref{extend_tuples} is the difference in scores across various similarity functions. \srl\ achieves generally higher correlation when using vector (spaCy) or ROUGE similarity over exact matching, although not to a significant degree. These observations can be attributed to the hypothesis that abstractive entity references will not be detected by exact matching. Also note that results on QAGS-XSUM are particularly affected by this, which shows higher levels of abstraction than CNN/DM-derived resources~\citep{wang-etal-2020-asking, pagnoni-etal-2021-understanding}. This is also visible for the SRLScore\textsubscript{coref} variant, as seen in \Cref{compare_SOTA}, which can further improve the matching of re-formulations.

\paragraph{Dynamic Weight Re-Normalization.}
\label{sec:dynamic}
We next analyze the contribution of our dynamic weighting scheme through removing the weight re-normalization $W_{norm}$ and instead defaulting to a static weighting on SRLScore\textsubscript{base}. Results in Table \ref{comparison-dynamicweights} demonstrate that re-distributing static weights dynamically to present roles is very effective, however, results show no statistical significance.

\begin{table}[t]
    \centering
    \begin{tabular}{p{0.29\linewidth}  p{0.05\linewidth}  p{0.05\linewidth}  p{0.05\linewidth} p{0.05\linewidth}p{0.05\linewidth} p{0.05\linewidth}} \toprule
	{\multirow{2}{*}{\centering \textbf{\footnotesize{Weight Setting}}}} & \multicolumn{2}{l}{\centering \textbf{\footnotesize{QCNNDM}}} & \multicolumn{2}{l}{\centering \textbf{\footnotesize{QXSUM}}} & \multicolumn{2}{l}{\centering \textbf{\footnotesize{SummE}}}  \\  \addlinespace[-3pt]
	\cmidrule(lr){2-3} \cmidrule(lr){4-5} \cmidrule(lr){6-7}
	 & \hspace*{0.5em}\textit{\footnotesize{$\rho$}} & \hspace*{0.5em}\textit{\footnotesize{s}}& \hspace*{0.5em}\textit{\footnotesize{$\rho$}}& \hspace*{0.5em}\textit{\footnotesize{s}}& \hspace*{0.5em}\textit{\footnotesize{$\rho$}}& \hspace*{0.5em}\textit{\footnotesize{s}}\\  \addlinespace[-3pt]                                                                                          

\midrule
{\centering \footnotesize{{Static weights}}} &\footnotesize{0.59}&\footnotesize{0.49}&\footnotesize{0.09}&\footnotesize{0.09}&\footnotesize{0.38}&\footnotesize{0.28} \\  \addlinespace[-3pt]

\midrule
{\centering \footnotesize{{Dynamic weights}}} & \footnotesize{\textbf{0.67}} &\footnotesize{\textbf{0.59}}     & \footnotesize{\textbf{0.20}}              &\footnotesize{\textbf{0.18}}           & \footnotesize{\textbf{{0.43}}} &\footnotesize{\textbf{0.33}}\\   \addlinespace[-3pt]
                                                                                                         
\bottomrule
    \end{tabular}
    \caption{Correlation scores of SRLScore\textsubscript{base} with and without weight re-normalization enabled.}
    \label{comparison-dynamicweights}
\end{table}

\paragraph{Ablation of Goodrich Scoring Method.}

We finally examine the performance of our scoring system against the partial matching approach of Goodrich et al. For fairness, we compare results on the reduced triplet sets. {SRLScore\textsubscript{\sloppy{openie}}} uses the presented weighting function, {SRLScore\textsubscript{\sloppy{goodrich}}} implements an equivalent scoring to Goodrich et al. Results in \Cref{comparison-method-effective} show that the presented scoring algorithm performs better than the scores determined by Goodrich's approach on different datasets, in most instances to a significant degree.

\begin{table}[t]
    \centering
    \begin{tabular}{p{0.31\linewidth}  p{0.048\linewidth}  p{0.048\linewidth}  p{0.048\linewidth} p{0.048\linewidth}p{0.048\linewidth} p{0.048\linewidth}} \toprule
	{\multirow{2}{*}{\centering \textbf{\footnotesize{Scoring Method}}}} & \multicolumn{2}{l}{\centering \textbf{\footnotesize{QCNNDM}}} & \multicolumn{2}{l}{\centering \textbf{\footnotesize{QXSUM}}} & \multicolumn{2}{l}{\centering \textbf{\footnotesize{SummE}}}  \\  \addlinespace[-3pt]	
	\cmidrule(lr){2-3} \cmidrule(lr){4-5} \cmidrule(lr){6-7}
	 & \hspace*{0.5em}\textit{\footnotesize{$\rho$}} & \hspace*{0.5em}\textit{\footnotesize{s}}& \hspace*{0.5em}\textit{\footnotesize{$\rho$}}& \hspace*{0.5em}\textit{\footnotesize{s}}& \hspace*{0.5em}\textit{\footnotesize{$\rho$}}& \hspace*{0.5em}\textit{\footnotesize{s}}\\   \addlinespace[-3pt]
\midrule
{\footnotesize{SRLScore\textsubscript{goodrich}}} &\footnotesize{0.45}&\footnotesize{0.38}&\footnotesize{0.05}&\footnotesize{0.07}&\footnotesize{0.29}&\footnotesize{0.24}\\   \addlinespace[-3pt]

\midrule
{\footnotesize{{SRLScore\textsubscript{openie}}}}   &\footnotesize{\textbf{0.62}\textsuperscript{\dag}} &\footnotesize{\textbf{0.56}\textsuperscript{\dag}}                   &\footnotesize{\textbf{0.13}} &\footnotesize{\textbf{0.10}}         &\footnotesize{\textbf{{0.41}}\textsuperscript{\ddag}} &\footnotesize{\textbf{0.32}\textsuperscript{\dag}}\\   \addlinespace[-3pt]
                                                                                                     
\bottomrule
    \end{tabular}
    \caption{Results of the ablation experiment comparing the scoring method by \citet{goodrich-etal-2019-assesssing} with our proposed scheme, based on triplet representations.}
    \label{comparison-method-effective}
\end{table}

\paragraph{Performance of Co-reference Resolution System.}

Results in \Cref{compare_SOTA} reveal that the co-reference system is not always improving scores, particularly on the CNN/DailyMail-derived datasets. However, the use of co-reference resolution will significantly increase the processing time, as shown in Table \ref{runtime}. This is expected, given that there are now more fact tuples due to the \textit{tuple expansion}; since the presented scoring method requires the comparison of each fact tuple in the summary against \emph{all} input text tuples.
We further compare the runtime against BARTScore, which only requires a single forward-pass through a neural net and can be batched easily, resulting in a 10x speed-up. In contrast, \srl\ requires construction and comparison the fact tuples, which are the main contributors for slower inference times.

\subsection{Error Analysis}

\begin{table}[t]
	\centering
	\begin{tabular}{p{0.12\linewidth} p{0.12\linewidth}  p{0.12\linewidth} p{0.12\linewidth} p{0.15\linewidth}} \toprule
		\multicolumn{2}{c}{\textbf{\footnotesize{SRLScore}}} & \multicolumn{3}{c}{\textbf{\footnotesize{BARTScore}}}\\ \addlinespace[-3pt]
		
		\cmidrule(lr){1-2} \cmidrule(lr){3-5} 
		
		\footnotesize{base} & \footnotesize{coref} & \footnotesize{base} & \footnotesize{cnn} & \footnotesize{cnn+para}\\ \addlinespace[-3pt]
		
		\midrule
		\footnotesize{2.35}  &\footnotesize{19.32}&\footnotesize{0.22}&\footnotesize{0.23}&\footnotesize{0.23}\\  \addlinespace[-3pt]
		
		\bottomrule
	\end{tabular}
	\caption{Average processing time (in seconds) per instance in QAGS-CNN/DM. \srl\ uses ROUGE similarity. BARTScore is run with a batch size of 4.}
	\label{runtime}
\end{table}

To better understand the limitations of our presented methods, we examine a number of instances manually, particularly those where there are large differences between model-generated scores and human annotations on QAGS-XSUM.
\Cref{manual} shows two instances, where \srl\ respectively predicts a much higher and lower factuality score than human annotators. 
Notably, human raters tend to drastically reduce factuality scores in the presence of even a single mistake (what we refer to as \emph{"strike-out scoring"}). In comparison, \srl\ and other factuality metrics tend to be more heavily influenced by the correctness of the \emph{majority} of attributes, which can be seen as a \emph{"bottom-up scoring"} (scores are built up from a initial factuality of zero instead of deducing from an initial score of one). 
On the other hand, highly abstractive samples, which retain factuality according to human raters, may pose a challenge for tuple-based \srl.
In the second example of \Cref{manual}, synonymous expressions like \emph{step down} instead of \emph{resign} cause low predicted similarity; potential solutions could be found in verb sense disambiguation~\citep{brown-etal-2011-verbnet, brown-etal-2022-semantic}.

\begin{table*}[t]
	\centering
	\begin{tabular}{p{0.07\linewidth} p{0.31\linewidth}  p{0.36\linewidth} p{0.045\linewidth} p{0.09\linewidth}} \toprule
		\footnotesize{} & \footnotesize{\textbf{Sample Text}} & \footnotesize{\textbf{Extracted Fact Tuples}} & \footnotesize{\textbf{Human}} & \footnotesize{\textbf{SRLScore}}\\ 
		\footnotesize{Input} & \footnotesize{\textcolor{ForestGreen}{Former England fast bowler Chris Tremlett} has announced his retirement ...} & \fontsize{7.8}{9}\selectfont{(\textcolor{ForestGreen}{\texttt{Former England fast bowler chris tremlett}}, \texttt{announce}, \texttt{his retirement}, ...}) & \multirow{2}{*}{\hspace*{1em}\footnotesize{0}} & \multirow{2}{*}{\hspace*{1em}\footnotesize{0.87}}\\ \addlinespace[-3pt]
		
		\footnotesize{Summary} &   \footnotesize{\textcolor{red}{Former England seamer James Tremlett} has announced his retirement ...}  & \fontsize{7.8}{9}\selectfont{(\textcolor{red}{\texttt{Former England seamer james tremlett}}, \texttt{announce}, \texttt{his retirement}, ...)} \\ \addlinespace[-3pt]
		\midrule
		\footnotesize{Input} & \footnotesize{The head of Japanese advertising group Dentsu is to \textcolor{ForestGreen}{step down} following \textcolor{ForestGreen}{the suicide of an employee} ...} & \fontsize{7.8}{9}\selectfont{(\texttt{The head of japanese advertising group dentsu}, \texttt{\textcolor{ForestGreen}{step}}, ..., \texttt{following \textcolor{ForestGreen}{the suicide of an employee}},  ...)} & \multirow{2}{*}{\hspace*{1em}\footnotesize{1}} & \multirow{2}{*}{\hspace*{1em}\footnotesize{0.10}}\\ 
		\footnotesize{Summary} & \footnotesize{The chief executive of Japanese advertising firm Dentsu will \textcolor{ForestGreen}{resign} after \textcolor{ForestGreen}{a worker killed herself} ...} &\fontsize{7.8}{9}\selectfont{(\texttt{The chief executive of japanese advertising firm dentsu}, \texttt{\textcolor{ForestGreen}{resign}}, ..., \texttt{after \textcolor{ForestGreen}{a worker killed herself}}, ...), (\texttt{a worker}, \texttt{killed}, \texttt{herself}, ...)} \\  \addlinespace[-3pt]
		
		\bottomrule
	\end{tabular}
	\caption{Examples from the QAGS-XSUM dataset where the majority vote of human ratings differs strongly from \srl's predicted factuality. Colored text segments highlight the position of relevant facts, where red  text indicates a factual discrepancy between input and summary segments.}
	\label{manual}
\end{table*}

\section{Conclusion and Future Directions}

In this work, we presented a semantically consistent metric for estimating the factual truthfulness of two pieces of text: we applied our presented metric to the problem of text summarization evaluation, and demonstrated that it performs on par with existing approaches.
In fact, we find that due to the small sample sizes of evaluation datasets, there are no significant differences between any of the considered state-of-the-art factuality estimation metrics.
Our approach strikes with its relative simplicity and interpretability due to the intermediate representation of "fact tuples", which makes it possible for human annotators to review how or why system decisions were made.
Furthermore, we have demonstrated the suitability of our approach over more naive tuple-based scoring methods through a series of ablation experiments, which also show the adaptability of our method to particular unseen settings by simply adjusting a series of parameters.\\

In our opinion, there are two key challenges concerning the effective deployment of \srl.
The current implementation still suffers from impractically long runtimes for longer input texts. Notably, however, both the tuple generation and comparison stages can be parallelized and we are currently working on improving the compute efficiency of our method.
Secondly, we have seen a general trend that factuality estimation metrics are scoring differently from human annotators, who are putting heavy emphasis on a \emph{completely} factual summary instead.
We suspect that adopting a similar \emph{strike-out scoring} for estimation may better correlate with human ratings, although it will require sufficiently accurate taggers to ensure correct recognition of all entities.

\section*{Limitations}
While the presented method exhibits stable correlation with human judgments on some of the evaluated datasets, it still exhibits instances under which it will predict opposing factuality scores. It should therefore be considered an \emph{addition} to human evaluation, but at this point not fully replace it.\\
We also want to point out that the underlying summarization datasets that were used to compare human ratings on are known for their own set of limitations, particularly being fairly extractive in nature.
This plays well with \srl's estimation of matching between individual tuples extracted from single sentences; on the other hand, if summary texts contain facts derived from multiple source sentences (or undergo otherwise complex structural changes), fact tuples may be insufficient in their current form.\\
Another limitation is the expressiveness of results on the fairly small human-annotated datasets. Here, statistically significant differences can rarely be obtained. However, we are to our knowledge the first to demonstrate this insight about (significant) differences between existing methods, which we consider a particularly useful insight for future work.

We further want to point out that our method was only evaluated on English datasets; we argue that it can be applied to other languages, given a similarly performing SRL labeling model. In practice, however, the existence of available models is currently limited for non-English languages.

\section*{Ethics Statement}

The paper considers the automated analysis of factuality in generated text.
While we see no imminent risk in the development of our presented method, we want to point to the explicitly spelled out limitations of the current method (see the previous section). The blind application of factuality metrics could be considered  harmful in instances where the predicted scores are differing strongly from human ratings. We therefore recommend that factuality metrics should be employed purely as a \emph{complementary} evaluation, and never directly replace analysis with humans in the loop.

\section*{Acknowledgments}
We thank the anonymous reviewers for their helpful comments and suggestions.
The work of Jing Fan is supported by a scholarship of the China Scholarship Council (CSC).

\bibliography{anthology,custom}
\bibliographystyle{acl_natbib}

\appendix
\section{Mapping of PropBank Arguments to Semantic Role Tuple Attributes}
\label{app:mapping}

In our implementation, we extract sentence spans with label ARG0 as \textit{agent} and spans with label ARG1 as \textit{patient}. The extraction of \textit{time} and \textit{location} also does not pose any difficulties, because ARGM-TMP and ARGM-LOC are both given as modifiers that remain relatively stable across predicates \citep{DBLP:books/lib/JurafskyM09}. However, as shown in Table \ref{ARG}, there is no one-to-one relationship between numbered arguments and the \textit{recipient} role. For the sake of simplicity, we extracted elements with label ARG2 as \textit{recipient}, because the probability that ARG2 correlates to \textit{recipient} is the highest among all other possible roles \citep{yi-etal-2007-can}.

\begin{table}[ht]
	\centering
	\begin{tabular}{p{0.08\linewidth} p{0.28\linewidth}  p{0.08\linewidth}  p{0.34\linewidth} }\toprule
		\textbf{\footnotesize{ARG0}} & \footnotesize{agent} & \textbf{\footnotesize{ARG1}} & \footnotesize{patient} \\  \addlinespace[-3pt]
		\midrule
		\textbf{\footnotesize{ARG2}} &\footnotesize{instrument, recipient, attribute} & \textbf{\footnotesize{ARG3}} &\footnotesize{starting point, recipient, attribute}  \\  \addlinespace[-3pt]
		\midrule
		\textbf{\footnotesize{ARG4}} & \footnotesize{ending point} & \textbf{\footnotesize{ARGM}} & \footnotesize{modifier}\\  \addlinespace[-3pt]
		\bottomrule
	\end{tabular}
	\caption{Mapping between numbered arguments in PropBank and semantic roles \citep{propbank}. Particularly the mapping of argument 2 makes simplifying assumptions about different verb forms.}
	\label{ARG}
\end{table}

\section{Reproducing Scores of Related Work}
\label{app:coco}
We use the official scripts provided by the authors of BARTScore\footnote{\url{https://github.com/neulab/BARTScore}, last accessed: 2023-02-01.} and CoCo\footnote{\url{https://github.com/xieyxclack/factual_coco}, last accessed: 2023-03-16.}.
Unfortunately, no public implementation exists at the time of writing for the work of \citet{li-etal-2022-just}, which prevents significance testing against ClozE models.
For the work by \citep{goodrich-etal-2019-assesssing}, we similarly found no publicly available implementation; however, we note their wikipedia-based training data for generating fact extractors is available online\footnote{\url{https://github.com/google-research-datasets/wikifact}, last accessed: 2023-05-17}.

When attempting to reproduce the scores of \citet{xie-etal-2021-factual-consistency}, based on their own implementation, we encountered wildly differing scores compared to the values reported by the authors.
Some results show drastic improvements from a reported Pearson correlation 0.58 to a reproduced score of 0.68, while other values dropped (e.g., on QAGS-XSUM, we see a reduction of scores from 0.24 to 0.16 in terms of Pearson correlation).
For the sake of reproducibility, we have included the exact commands that were used to run the CoCo models in our repository.

On the other hand, all of our reproduced scores for BARTScore~\cite{yuan-etal-2021-bartscore} match the available self-reported results by the authors.

For significance testing, we use our own implementation of a permutation-based significance test, again included in the code repository. We fix the initial \texttt{NumPy} random seed to \texttt{256}, and compute results over 10{,}000 iterations for each test.

\end{document}